\documentclass{ifacconf}

\usepackage{graphicx}
\usepackage{natbib}        
\usepackage{url}

\usepackage[subrefformat=parens]{subcaption}

\usepackage{amsmath}
\usepackage{amssymb}

\begin{document}
\begin{frontmatter}

\title{Versatile Telescopic-Wheeled-Legged Locomotion of Tachyon 3 via Full-Centroidal Nonlinear Model Predictive Control} 

\author[First]{Sotaro Katayama} 
\author[First]{Noriaki Takasugi} 
\author[Second]{Mitsuhisa Kaneko} 
\author[First]{Masaya Kinoshita} 
\address[First]{Sony Group Corporation, Minato-ku, Tokyo, Japan, 108-0075 \\ (email: sotaro.katayama@sony.com).}
\address[Second]{Sony Global Manufacturing \& Operations Corporation, Minato-ku, Tokyo, Japan, 108-0075}

\begin{abstract}
This paper presents a nonlinear model predictive control (NMPC) toward versatile motion generation for the telescopic-wheeled-legged robot Tachyon 3, the unique hardware structure of which poses challenges in control and motion planning.
We apply the full-centroidal NMPC formulation with dedicated constraints that can capture the accurate kinematics and dynamics of Tachyon 3.
We have developed a control pipeline that includes an internal state integrator to apply NMPC to Tachyon 3, the actuators of which employ high-gain position-controllers.
We conducted simulation and hardware experiments on the perceptive locomotion of Tachyon 3 over structured terrains and demonstrated that the proposed method can achieve smooth and dynamic motion generation under harsh physical and environmental constraints.
\end{abstract}


\end{frontmatter}

\section{Introduction}

Legged robots are promising robotic mobilities that can traverse a variety of places.
While industrial applications of legged robots, particularly quadruped robots, are nowadays gaining attention, typical legged robots still have two unresolved issues: energy efficiency and safety.
Legged robots are typically not energy efficient; they can work only less than an hour because they have to consistently support the entire mass of the robot using actuator power.
Moreover, they can lack safety, e.g., cause catastrophic damage to the robot itself and to the environment when they fall.

\begin{figure}[t]
  \centering
  \includegraphics[scale=0.45]{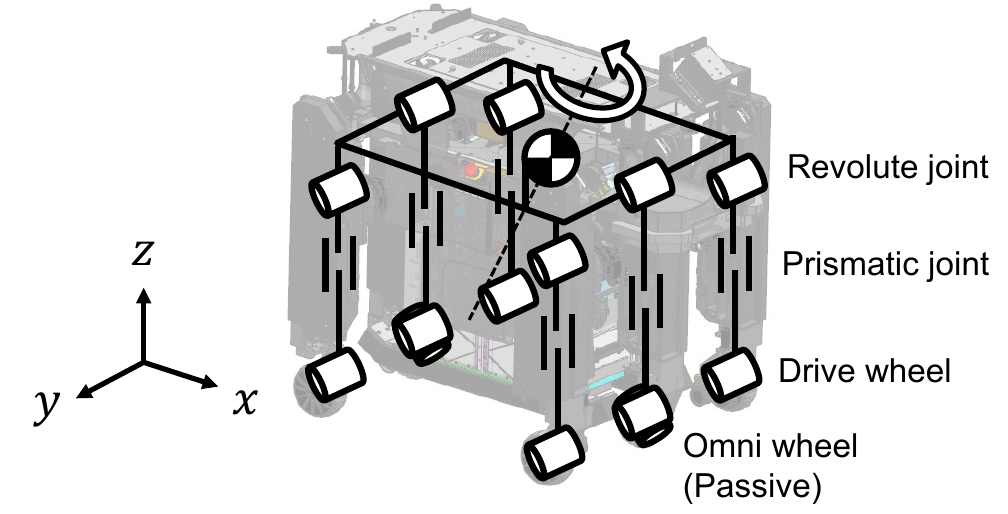}
  \caption{Hardware structure of telescopic-wheeled-legged robot Tachyon 3. It has three front legs (left-front (LF), middle-front (MF), and right-front (RF)) and three rear legs (left-rear (LR), middle -rear (MR) and right-rear (RR)). In total, it consists of 16 active joints including 6 prismatic joints, 6 hip joints and 4 drive wheels, and additional two passive omni wheels.}
  \label{fig:Tachyon3}
\end{figure}

To mitigate such issues and expand the capabilities of legged mobilities, we have developed a six-telescopic-wheeled-legged robot called Tachyon 3\footnote[1]{Website: \url{https://www.sony.com/en/SonyInfo/research/technologies/new_mobility}}\footnote[2]{Video: \url{https://youtu.be/sorw7o73ydc}}, which is shown in Fig. \ref{fig:Tachyon3}.
Each leg of Tachyon 3 is composed of a revolute joint at the hip, a telescopic joint at the knee, and a drive/passive wheel at the tip.
Remarkably, the knee joint is designed so that it can support the entire mass of the body without energy consumption at a particular joint position.
By combining the legged traversability and the wheel movement leveraging the knee joint structure, Tachyon 3 is expected to be an energy-efficient and versatile legged mobility.
Furthermore, its center of mass (COM) is designed to be located at a much lower position than typical legged robots, which can reduce damage due to accidents.

Aside from these advantages, the novel hardware structure of Tachyon 3 poses challenges in motion planning and control.
We summarize such challenges from the viewpoint of a comparison between typical quadruped robots.
\begin{itemize}
   \item Each foot of Tachyon 3 has only 2 degrees of freedom (DOF) relative to the body (x and z directions) while that of typical quadruped robots has 3 (x, y, and z directions).
   \item In addition to the previous point, LF, LR, RF, and RR feet have drive wheels and hence the whole system involves complicated non-holonomic constraints while typical quadruped robots only involve mathematically tractable point contacts.
   \item The joint ranges of Tachyon 3 are small (e.g., 50 degrees in MF and MR hip joints and 90 degrees in the other hip joints) to avoid self-collisions while quadruped robots have enough joint ranges (e.g., 180 degrees).
   \item Knee joints are required to operate near their position limit to keep the COM position lower, while typical quadruped robots are designed to have enough joint margin in the nominal configuration.
   \item Each joint of Tachyon 3 is controlled by a high-gain position-control while typical quadruped robots employ torque control.
\end{itemize}

To achieve safe motion generation in real time under the aforementioned hardware limitations, in \cite{takasugi2023realtime}, we have developed a safety-aware perceptive controller using a control barrier function (CBF) and quadratic programming (QP) controller (i.e., CBF-QP) that can run at 1 kHz.
However, the method alone can only take into account the instantaneous and myopic motions under a static stability derived from the kinematics.
As a result, the generated motion lacks smoothness, which often causes undesirable body oscillation and discontinuous body velocities in ascending steps.

To further expand the dynamic and smooth motion ability of Tachyon 3, in this paper, we have developed a nonlinear model predictive control (NMPC) framework for Tachyon 3.
Our NMPC is basd on the full-centroidal NMPC formulation (\cite{sleiman2021unified}) to capture the limited foot-wise DOFs, complicated non-holonomic constraints, and harsh kinematics and environmental constraints of Tachyon 3 in a unified manner.
We also ensure the safety by hierarchically combining NMPC and the CBF-QP controller (\cite{takasugi2023realtime}).
Further, to apply NMPC to Tachyon 3, the joints of which employ a high-gain position-control, we propose the use of an internal state integrator for state feedback of NMPC so that we can avoid oscillation in the resultant joint position commands.
We then integrated them into a control pipeline to achieve NMPC on the on-board computer of Tachyon 3.

\subsection{Related Works}
NMPC has been widely applied to legged robots, particularly for torque-controlled quadruped robot.
\cite{di2018dynamic} models the dynamics of a quadruped robot as single rigid body dynamics (SRBD) while the kinematics are inaccruately modeled.
\cite{farshidian2017efficient} and \cite{grandia2023perceptive} combine the SRBD with whole-body kinematics and further improve the accuracy in terms of the kinematics.
\cite{bjelonic2021whole,bjelonic2022offline} apply the NMPC formulation of \cite{farshidian2017efficient} to wheeled quadruped robot.
\cite{sleiman2021unified} improves the approach using whole-body kinematics and centroidal dynamics (\cite{orin2013centroidal}) instead of the SRBD to further improve accuracy.
These methods, called as reduced-order model-based NMPC, only computes optimal contact forces and kinematic quantities.
Therefore, for torque-controlled robots, a whole-body dynamics-based controller is typically employed to compute the joint torque commands from the NMPC solution.
By contrast, \cite{mastalli2022agile} and \cite{katayama2022whole} incorporate the whole-body dynamics into NMPC.
One can then improve the dynamics accuracy and also avoid using the additional whole-body dynamics-based controller at the expense of larger computational burden.

In the predictive controls of position-controlled robots such as humanoid robots, the SRBD, centroidal, and whole-body dynamics models have have garnered less attension that highly simplified models such as the linear inverted pendulum model (\cite{kajita2003biped}).
A pioneer study by \cite{koenemann2015whole} reports the applications of the whole-body dynamics-based NMPC for a position-controlled humanoid robot.
However, to avoid discontinuities in the joint position commands, it does not use any state feedback, that is, it is almost the same as an application of offline trajectory optimization.
Therefore, the resultant trajectory can be far from optimal or infeasible once the state prediction is different from the actual state.

\subsection{Contributions}
In this paper, we have developed NMPC for telescopic-wheeled-legged Tachyon 3 that has a novel and unique hardware structure.
Our contributions are summarized as follows:
\begin{itemize}
    \item Application of the full-centroidal NMPC (\cite{sleiman2021unified}) to the novel hardware Tachyon 3 with novel physical and environmental constraints.
    \item State feedback using an internal state integrator to apply NMPC to robots, the actuators of which employ high-gain position-controllers.
    \item Control pipeline and implementation details to integrate NMPC and other modules such as the state integrator and CBF-QP on an on-board PC that has limited computation resources.
    \item Validation through a simulation study against our previous controller (\cite{takasugi2023realtime}) and real-world hardware experiment on perceptive locomotion 
\end{itemize}

This paper is organized as follows.
Section \ref{Sec:MPC} proposes the NMPC formulation of Tachyon 3.
Section \ref{Sec:system} introduces implementation of NMPC from the viewpoint of a numerical optimization solver to the overall system architecture implemented on the on-board PCs of Tachyon 3.
Section \ref{Sec:experiments} shows the effectiveness of the proposed NMPC framework through simulation and hardware experiments on the perceptive locomotion over structured terrains.
Finally, Section \ref{Sec:Conclusion} concludes this paper with future works.

\section{Nonlinear Model Predictive Control of Tachyon 3}\label{Sec:MPC}
\subsection{Optimal Control Problem of Switched Systems}
In robotic applications, NMPC is often referred to as an online motion planning method that solves an optimal control problem (OCP) each time.
We herein formulate an OCP for Tachyon 3 under a fixed contact sequence, i.e., a sequence of contact flags (on/off) of all feet and the switching times.
We define a mode for each combination of contact flags of six feet.
An OCP under such a setting can be defined as a general OCP of switched systems under a given mode sequence $[m_1, ..., m_M]$ and switching times $[t_0, ..., t_{M-1}]$, where $t_i$ denotes the switching time from mode $m_i$ to $m_{i + 1}$.
Then, for the given horizon $[t_0, t_M]$ and initial state $\bar{x}$, the OCP is formulated as
\begin{align}
    & \min_{u(\cdot)} V_{m_M} (t_M, x(t_M)) + \sum_{m=0}^{m=M-1} \int_{t_m}^{t_{m+1}} l_m (t, x (t), u (t)) \notag \\
    & {\rm s.t.} \;\;\; x(t_0) = \bar{x}, \; \dot{x} (t) = f_m (t, x(t), u(t)) \;\; (t_m \leq t < t_{m+1}) \notag \\
    & \;\;\;\;\;\;\;\; e_m (t, x(t), u(t)) = 0 \;\; (t_m \leq t < t_{m+1}) \notag \\
    & \;\;\;\;\;\;\;\; g_{m} (t, x(t), u(t)) \geq 0 \;\; (t_m \leq t < t_{m+1}) \notag \\
    & \;\;\;\;\;\;\;\; g_M (t_M, x(t_M)) \geq 0.
\end{align}
Here, subscript $m$ indicates that a function is given for the mode $m$ and $V_{m} (\cdot)$, $l_{m} (\cdot)$, $f_m (\cdot)$, $e_m (\cdot)$ and $g_m (\cdot)$ denote the terminal cost, stage cost, state equation, equality constraint, and inequality constraint, respectively.

In NMPC, at each time $t$, we measure or estimate the current state $x(t)$ and solves the above OCP with $t_0 = t$, $\bar{x} = x(t)$, and $t_M = t + T$, where $T > 0$ is the horizon length.
The latest optimal trajectory is then utilized to determine the control action of the system.

\subsection{Full-Centroidal Model}\label{Sec:Model}
We model Tachyon 3 using the full-kinematics and centroidal dynamics model as is the case with \cite{sleiman2021unified}, which is termed a \textit{full-centroidal model}. 
The full-kinematics can capture the precise constraint while the centroidal dynamics can express the accurate dynamics under certain assumptions (\cite{orin2013centroidal}).
The precise constraint evaluation is particularly crucial for Tachyon 3, which has severe joint range limits, limited foot-wise DOF, and non-holonomic wheel contact constraints.
The full-centroidal model has a good balance between accuracy and computational cost; for example, it is much computationally cheaper than the exact whole-body dynamics formulation.

The state vector of the full-centroidal model is defined as 
\begin{equation}
    x
    := \begin{bmatrix}
        q_{\rm b} ^{\rm T} &
        q_{\rm J} ^{\rm T} &
        h_{\rm com} ^{\rm T}
    \end{bmatrix} ^{\rm T}
    := 
    \begin{bmatrix}
        q ^{\rm T} &
        h_{\rm com} ^{\rm T}
    \end{bmatrix}^{\rm T} , 
\end{equation}
where $q_{\rm b} \in SE(3)$ denotes the base pose (position and rotation), $q_{\rm J} \in \mathbb{R}^{n_{\rm J}}$ denotes the joint positions ($n_{\rm J} = 12$ in the case of Tachyon 3), $q$ denotes the configuration, and $h_{\rm com} \in \mathbb{R}^6$ denotes the centroidal momentum.
We denote the time variation of the configuration $q$ at its tangent space as 
\begin{equation}
    v := 
    \begin{bmatrix}
        v_{\rm b} ^{\rm T} &
        v_{\rm J} ^{\rm T}
    \end{bmatrix} ^{\rm T} 
    = \begin{bmatrix}
        v_{\rm b} ^{\rm T} &
        \dot{q}_{\rm J} ^{\rm T}
    \end{bmatrix} ^{\rm T} ,
\end{equation}
where $v_{\rm b} \in \mathbb{R}^6$ denotes the spatial (linear and angular) velocity of the base expressed in the base local coordinate frame.
The control input of the model is defined as 
\begin{equation}
    u := 
    \begin{bmatrix}
        v_{\rm J} ^{\rm T} & 
        f_1 ^{\rm T} &
        \cdots &
        f_{n_c} ^{\rm T}
    \end{bmatrix} ^{\rm T},
\end{equation}
where $f_i \in \mathbb{R}^3$ denotes the contact force acting on the $i$-th foot and $n_c$ is the number of feet ($n_c = 6$ in the case of Tachyon 3).

To derive the state equation, we introduce the centroidal momentum matrix $ A(q) \in \mathbb{R}^{6 \times (6 + n_{\rm J})}$, which maps the generalized velocity onto the centroidal momentum (\cite{orin2013centroidal}) as 
\begin{equation}
    h_{\rm com} = 
    A(q)
    \begin{bmatrix}
        v_{\rm b} \\
        v_{\rm J}
    \end{bmatrix}
    = \begin{bmatrix}
        A_{\rm b} (q) &
        A_{\rm J} (q)
    \end{bmatrix}
    \begin{bmatrix}
        v_{\rm b} \\
        v_{\rm J} 
    \end{bmatrix}, 
\end{equation}
where $A_{\rm b} (q) \in \mathbb{R}^{6 \times 6}$ and $A_{\rm b} (q) \in \mathbb{R}^{6 \times n_{\rm J}}$.
The state equation is then given by
\begin{equation}\label{eq:stateEquation}
    f(x, u) = 
    \begin{bmatrix}
        A_{\rm b} (q) ^{-1} (h_{\rm com} - A_{\rm J} (q) v_{\rm J}) \\
        v_{\rm J} \\
        \sum_{i=1}^{i_{n_c}} f_i + m g \\
        \sum_{i=1}^{i_{n_c}} ({\rm FK}_i (q) - {\rm COM} (q)) \times f_i
    \end{bmatrix} ,
\end{equation}
where $m$ is the total mass of the robot, ${\rm FK}_i (q) \in \mathbb{R}^3$ is a forward kinematics, i.e., position of the $i$-th foot, and ${\rm COM} (q) \in \mathbb{R}^3$ is the COM position.

\subsection{Equality Constraints}
\subsubsection{Contact constraints:}
We impose wheel contact constraints for a given contact schedule in a way similar to \cite{bjelonic2021whole,bjelonic2022offline} but extended to Tachyon 3.
A swing foot does not have any contact forces, hence an equality constraint
\begin{equation}
    f_i = 0  
\end{equation}
is imposed for each swing foot $i$.
A stance foot cannot make any motion in the z-direction.
Therefore, an equality constraint
\begin{equation}\label{eq:zeroWheelVelocityConstraint}
    \begin{bmatrix}
        0 & 0 & 1
    \end{bmatrix}
    {\rm v}_i (q, v) = 0
\end{equation}
is imposed for each stance foot $i$, where ${\rm v}_i (q, v) \in \mathbb{R}^3$ denotes the linear velocity of the $i$-th foot expressed in the world coordinate frame.

In addition to (\ref{eq:zeroWheelVelocityConstraint}), LF, LR, RF, and RR feet have drive wheels and therefore cannot have any motions in the local $y$-direction. 
We then consider an equality constraint
\begin{equation}\label{eq:passiveOmniWheelVelocityConstraint}
    \begin{bmatrix}
        0 & 1 & 0
    \end{bmatrix}
    R_{{\rm yaw}, i} ^{\rm T} (q)
    {\rm v}_i (q, v) = 0
\end{equation}
for each of LF, LR, RF, and RR contacts, where $R_{{\rm yaw}, i} (q)$ is a rotation matrix that extracts the yaw rotation of $i$-th foot.
However, we found that directly imposing (\ref{eq:zeroWheelVelocityConstraint}) and (\ref{eq:passiveOmniWheelVelocityConstraint}) for all stance feet in the case of Tachyon 3 can result in redundancy in the equality constraints and cause numerical optimization failures.
We hypothesize that imposing (\ref{eq:zeroWheelVelocityConstraint}) and (\ref{eq:passiveOmniWheelVelocityConstraint}) for two side-by-side feet, such as LF and RF feet, doubly constrains the angular motion of the body around the local $y$-axis because of the small footwise DOF of Tachyon 3.
From this perspective, we remove the equality constraints.
Specifically, we impose (\ref{eq:zeroWheelVelocityConstraint}) and (\ref{eq:passiveOmniWheelVelocityConstraint}) for only one of the two side-by-side feet when both of them have contacts.
Although this constraint switch is a heuristic, it works well in practice.

\subsubsection{Swing foot trajectory constraint:}
In addition to the contact constraints, we impose swing foot height trajectory as an equality constraint as \cite{farshidian2017efficient}.
Suppose that the heights of each contact are provided as well as the contact flags from a higher layer planner.
We then design the swing foot height trajectory $z(t) \in \mathbb{R}$ by spline curves for the given contact heights as 
\begin{equation}\label{eq:swingFootTrajectoryConstraint}
    \begin{bmatrix}
        0 & 0 & 1
    \end{bmatrix}
    {\rm v}_i (q, v) - \dot{z}_{i} (t) + 
    K_p 
    (\begin{bmatrix}
        0 & 0 & 1
    \end{bmatrix}
        {\rm FK}_i (q) - {z}_{i} (t)) 
    = 0, 
\end{equation}
where $K_p \geq 0$ is a position-error feedback gain.

\subsection{Inequality Constraint}\label{Sec:InequalityConstraint}

\subsubsection{Joint position, velocity, and torque limit constraints:} 
Thanks to the full-kinematics model, we can directly impose the position and velocity limit inequality constraints on each joint such as 
\begin{equation}
    q_{\rm J, min} \leq q_{\rm J} \leq q_{\rm J, max}, \;\;
    v_{\rm J, min} \leq v_{\rm J} \leq v_{\rm J, max}.
\end{equation}
We also impose static joint torque limit as 
\begin{equation}\label{eq:torqueLimit}
    \tau_{\rm J, min} \leq S (g(q) + J^{\rm T} (q) f) \leq \tau_{\rm J, max},
\end{equation}
where $S \in \mathbb{R}^{n_{\rm J} \times (6 + n_{\rm J})} $ denotes the selection matrix, $g(q) \in \mathbb{R}^{6 + n_{\rm J}}$ denotes the generalized gravity term, $J(q) \in \mathbb{R}^{3 n_c \times (6 + n_{\rm J})}$ denotes the stack of the contact Jacobians, and $f \in \mathbb{R}^{3 n_c}$ denotes the stack of the contact forces, respectively.

\subsubsection{Friction cone constraints:} 
To avoid negative normal force and slipping, we impose a linearized friction cone for each active contact as used in \cite{katayama2022whole}.

\subsubsection{Foot feasible region constraints:} 

\begin{figure}[tb]
    \centering
    \begin{minipage}{0.49 \linewidth}
        \centering
        \includegraphics[scale=0.36]{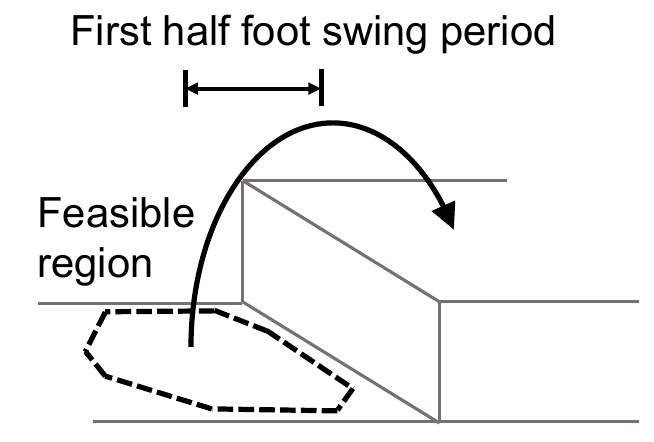}
        \subcaption{Ascending a step}
    \end{minipage}
    \begin{minipage}{0.49 \linewidth}
        \centering
        \includegraphics[scale=0.36]{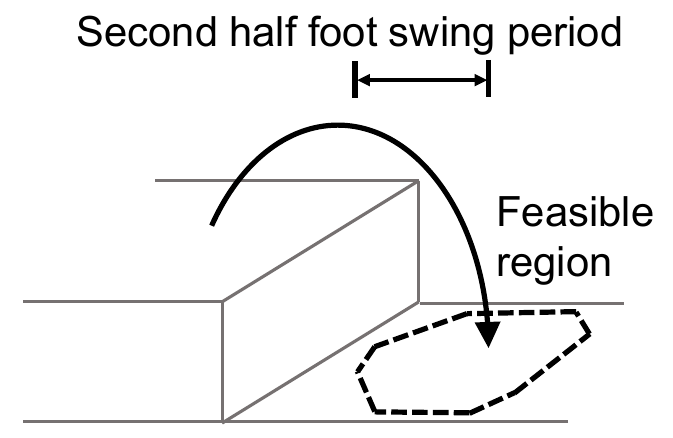}
        \subcaption{Descending a step}
    \end{minipage}
    \caption{Feasible regions for swing foot collision avoidance}
    \label{fig:feasibleStep}
\end{figure}

In walking over terrain, the foot position projected onto the $(x, y)$-plane should be constrained to avoid 1) falling from a terrain surface and 2) collisions between legs and the terrain.
To this end, we impose the $(x, y)$ position of the foot within a convex polygonal region extracted from the perception pipeline.
The convex polygon is expressed as 
\begin{equation}\label{eq:convexRegionConstraint}
    A 
    \begin{bmatrix}
        1 & 0 & 0 \\
        0 & 1 & 0 
    \end{bmatrix}
    {\rm FK}_i(q) + b \geq 0,
\end{equation}
where $A$ and $b$ are the coefficients of the polygon.
For each stance foot, the convex region constraint is imposed as well as the previous studies (\cite{grandia2023perceptive,bjelonic2022offline}).

In addition, we leveraged the convex region constraint (\ref{eq:convexRegionConstraint}) for swing foot collision avoidance.
This is in contrast to typical collision avoidance NMPC formulations based on signed distance fields such as \cite{grandia2023perceptive} with which high nonlinearity can cause numerical difficulties.
Fig. \ref{fig:feasibleStep} illustrates the proposed convex region constraint for the swing foot collision avoidance.
Suppose that the swing foot height trajectory in (\ref{eq:swingFootTrajectoryConstraint}) monotonically increases in the first-half of the foot swing period and decreases in the second-half of it.
Then, in the case of ascending a step, only the first-half swing foot period can have a collision.
We can therefore achieve swing foot collision avoidance by imposing (\ref{eq:convexRegionConstraint}) over the first-half swing foot period in the ascending case with an appropriate convex region.
Similarly, in the descending case, we impose (\ref{eq:convexRegionConstraint}) over the second-half of the swing foot period.

\subsection{Cost Function}\label{Sec:Cost}

\subsubsection{Base local velocity tracking cost:} 
As with typical legged robots, the move command to Tachyon 3 is the base velocity command expressed in the base local coordinate frame.
According to (\ref{eq:stateEquation}), the base velocity is expressed as a nonlinear function of $x$ and $u$:
\begin{equation}\label{eq:bodyVelocityMap}
    v_{\rm b} (x, u) := A_b (q) ^{-1} (h_{\rm com} - A_J (q) v_J).
\end{equation}
For given velocity command $v_{\rm ref}$, we consider a nonlinear least-square cost term the residual of which is given by 
$v_{\rm b} (x, u) - v_{\rm ref}$.

\subsubsection{State and control input tracking cost:} 
We also consider a quadratic state-input tracking cost to track $x_{\rm ref} (t)$ and $u_{\rm ref} (t)$.
Suppose that time-varying reference base pose $q_{\rm b, ref} (t)$ is provided by a higher-layer planner.
Suppose also that the nominal joint positions are given by constant vector $q_{\rm J, ref}$.
The reference state is then given by 
\begin{equation}
    x_{\rm ref} (t)
    = \begin{bmatrix}
        q_{\rm b, ref} (t) ^{\rm T} &
        q_{\rm J, ref} ^{\rm T} &
        0_{6 \times 1} ^{\rm T}
    \end{bmatrix}^{\rm T} .
\end{equation}
Each element of reference control input $u_{\rm ref} (t)$ is zero except for the $z$-element of $f_i$ of stance foot $i$; it is given by division of the total weight by the number of active contacts.

\section{Implementation}\label{Sec:system}

\begin{figure*}[tb]
  \centering
  \includegraphics[scale=0.42]{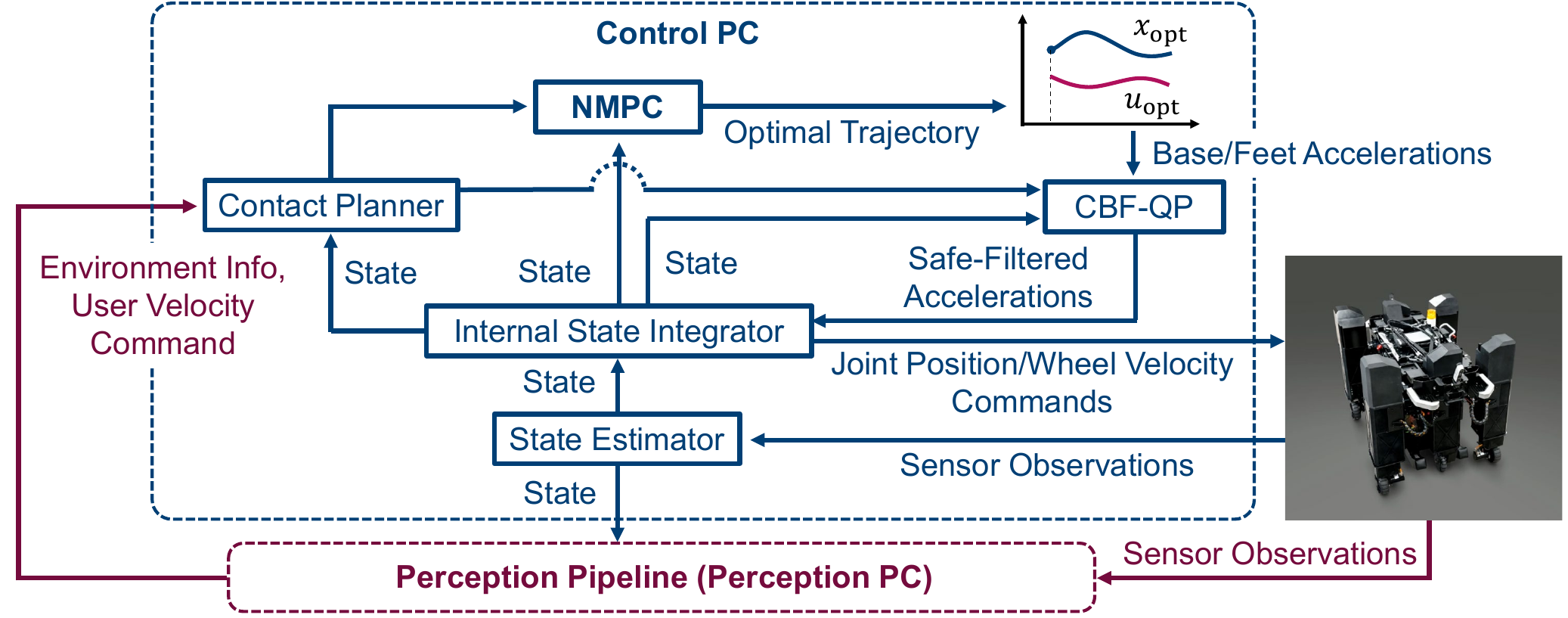}
  \caption{Control pipeline of Tachyon 3 including NMPC}
  \label{fig:systemArch}
\end{figure*}

\begin{figure}
  \centering
  \includegraphics[scale=0.45]{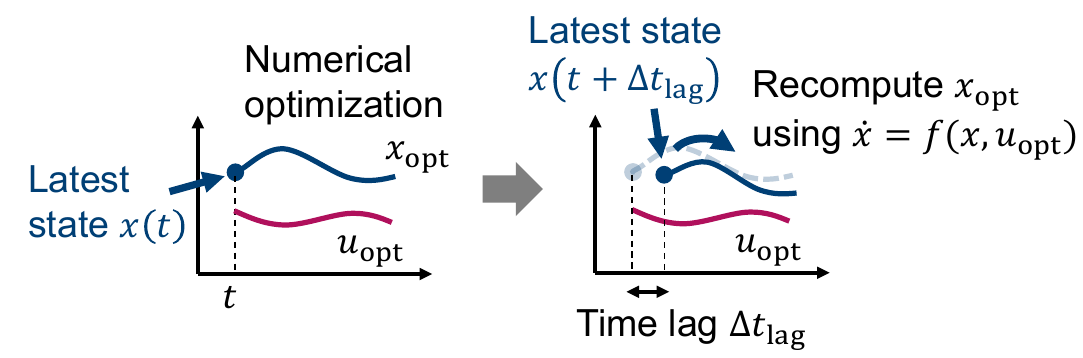}
  \caption{State trajectory refinement for time lag $\Delta t_{\rm lag}$.}
  \label{fig:statetrajrefine}
\end{figure}

\subsection{Numerical Optimization Solver}\label{Sec:solver}
The OCP of Tachyon 3 presented so far has many terms that depend on contact flags.
It is not easy to implement such problems efficiently using general-purpose nonlinear programming (NLP) solvers such as Ipopt (\cite{wachter2006implementation}), particularly to extract the problem-specific structure and warm-start strategy.
On the other hand, \texttt{OCS2} (\cite{OCS2Farbod}) is a promising open-source framework for such NMPC implementation and it has been utilized in NMPC for legged robots (\cite{farshidian2017efficient,sleiman2021unified,bjelonic2021whole,bjelonic2022offline,grandia2023perceptive}).
However, it is not suitable for embedded implementation as it deeply depends on Robot Operation Sysmte (ROS) and involves a large amount of runtime temporary memories.

To mitigate such issues, we have developed a numerical optimal control solver, which employs a similar numerical method and interface similar to that of \texttt{OCS2} while avoiding the aforementioned technical issues.
In our solver, the OCP is transcribed into a finite-dimensional NLP using the direct multiple shooting method (\cite{bock1984multiple}).
It employs the structure-exploiting primal-dual interior point method (\cite{nocedal1999numerical}) and Gauss-Newton Hessian approximation (\cite{rawlingsmodel}) to solve the NLP.
Equality constraints are handled using the projection method (\cite{nocedal1999numerical,farshidian2017efficient}).
The open-source structure-exploiting numerical solver HPIPM (\cite{frison2020hpipm}) is used to compute the search-direction efficiently. 
Further, our solver can directly handle Lie-groups within the direct multiple shooting method in contrast to \texttt{OCS2} that consistently depends on Euler-angles.
To implement algorithms regarding robot kinematics and dynamics, we use the C++ open-source library \texttt{Pinocchio} (\cite{carpentier2019pinocchio}).
In contrast to \texttt{OCS2} which consistently uses automatic differentiation for kinematics and dynamics, we utilize the analytical derivatives provided by \texttt{Pinocchio} (\cite{carpentier2018analytical}) to reduce the redundant computations (e.g., frame Jacobians $J(q)$ and body local velocity (\ref{eq:bodyVelocityMap})). 
\texttt{Pinocchio} also enables us to handle Lie-groups directly in the numerical optimization.
Thanks to the careful implementation and the efficient use of \texttt{Pinocchio}, our solver can improve the computation speed slightly over that of \texttt{OCS2}. 

\subsection{Real-Time NMPC}\label{Sec:nmpc}

In the following experiments on NMPC, the OCP solver performed Newton-type iteration ones each sampling time as was the case with \cite{diehl2005real}.
We fixed the barrier parameter of the primal-dual interior point method to enable fast computation.
We also used a soft constraint (\cite{feller2016relaxed}) for state-only inequality constraints with the fixed barrier and relaxation parameters  to avoid numerical difficulties.
We set the horizon length to 1.5 s.
We set the discretization time step for direct multiple shooting to 0.015 s (i.e., there were around 100 discretization grids over the horizon).

\subsection{Control Pipeline and Implementation Details}\label{Sec:control}

Fig. \ref{fig:systemArch} shows the overall control pipeline.
Tachyon 3 has two on-board PCs. One is a \textit{perception PC} and the other is a \textit{control PC}.
The perception PC is used for environment perception based on exteroceptive sensors such as LiDAR.
In particular, it extracts surfaces used in (\ref{eq:convexRegionConstraint}) in a way similar to \cite{grandia2023perceptive} as well as obstacles.
The control PC (CPU: Intel(R) Core(TM) i7-8850 H CPU @ 2.606 GHz) is used for motion planning and control based on the proprioceptive sensors and the environment information sent from the perception pipeline.
The main thread (real-time thread) of the control PC runs at 1 kHz and includes the contact planner, CBF-QP, internal state integrator, and state estimator.
For the state estimator, we used a variant of the extended Kalman filter (\cite{hartley2020contact}).
NMPC is implemented on another thread of the control PC.
The NMPC thread asynchronously receives state and contact plans from the real-time thread and sends the optimal trajectory to the real-time thread.
NMPC runs as fast as possible and we observed that it runs around 25 Hz.

The optimal state trajectory as well as the optimal control input trajectory are used to compute the desired acceleration.
However, numerical optimization causes a non-negligible time-lag and therefore the predicted optimal state trajectory can have some error from the actual state.
To compensate for the error from the actual state, after solving the NMPC optimization problem, we refined the optimal state trajectory based on the latest received state.
This is achieved by integrating the state over the horizon using the state equation (\ref{eq:stateEquation}) with the latest received state and the optimal control input as shown in Fig. \ref{fig:statetrajrefine}.

From the refined optimal trajectory, the positions, velocities, and accelerations of the base pose and feet are extracted for a given queried time.
The acceleration commands are then computed by the sum of feedforward and feedback terms.
For example, the acceleration command of the $i$-th foot is given by 
\begin{equation}
    a_{i} + K_p (p_i - \bar{p}_{i}) + K_d (v_i - \bar{v}_i)
\end{equation}
where $p_{i}, v_{i}, a_i \in \mathbb{R}^3$ are the position, velocity, and acceleration of the $i$-th foot extracted from the optimal trajectory, $\bar{p}_{i}, \bar{v}_{i} \in \mathbb{R}^3$ are the current velocity and position of the $i$-th foot retrieved from the internal state integrator, and $K_p, K_d \geq 0$ are the position and velocity feedback gains.
The acceleration commands are further filtered by the CBF-QP and the safe-filtered accelerations are then sent to the state integrator to update its internal state.
The CBF-QP is of practical importance for a rigorous safety because the update rate of NMPC is limited.

\begin{figure}
  \centering
  \includegraphics[scale=0.72]{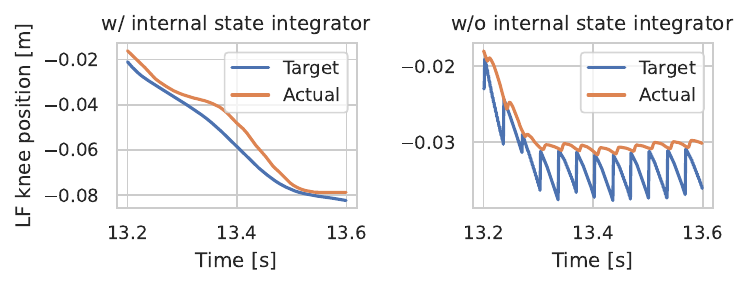}
  \caption{Time histories of target and actual joint positions measured in real robot experiment with the internal state integrator (left) and without it (right).
  The state integrator could successfully remove oscillation that appears in the case of the naive state feedback without using it.
  }
  \label{fig:sindou}
\end{figure}

The internal state integrator in Fig. \ref{fig:systemArch} also plays an essential role for Tachyon 3, which employs high-gain joint position-controllers.
As reported in \cite{koenemann2015whole}, NMPC for position-controlled robots with naive state feedback causes oscillation in the joint position commands due to the prediction error of NMPC, which is inevitable because the numerical optimization of NMPC has a non-negligible time lag.
For this issue, in the same spirit as \cite{kuindersma2016optimization}, we employ an internal state integrator module that updates the internal state of the robot by a combination of the output acceleration of the CBF-QP and the estimate of the actual state.
With the state integrator, we can avoid oscillation in the resultant joint position commands while consistently feedback the actual state. 
This is in contrast to previous NMPC for a position controlled robot (\cite{koenemann2015whole}) which does not use any actual state feedback.

We examined NMPC with and without the internal state integrator in real hardware.
Fig. \ref{fig:sindou} shows the time histories of target and actual joint positions of the experiments.
We can see the oscillation in target joint commands of NMPC using the naive state feedback without using the state integrator while NMPC using the state integrator resolved the issue.
We believe that this framework enables the application of NMPC based on the full-kinematics not only to Tachyon 3 but also to other position-control-based robots.

\section{Experiments}\label{Sec:experiments}
\subsection{Settings}\label{Sec:experimentsSettings}
To demonstrate the effectiveness of the proposed NMPC framework, we conducted simulation and hardware experiments.
We assumed realistic settings, that is, Tachyon 3 did not have any prior information regarding the environment and only relied on its on-board sensors. 
The proprioceptive sensor observations of Tachyon 3 were the joint positions from joint encoders, local base linear acceleration and angular velocity from the IMU. 
Also, the foot contact sensors and actuator current were used to detect foot contacts for the state estimation.
The exteroceptive sensor observations were point coulds from a LiDAR.
Throughout the experiments, we considered trot\footnote[3]{We define trot as a walking pattern alternating [LF, RF, MR] feet contact and [MF, LR, RR] feet contact.} gait.
Note that our NMPC is not limited to a trot: it can realize a variety of gait patterns as long as the numerical optimization converges.

\begin{figure}[tb]
    \centering
    \begin{minipage}{1.0 \linewidth}
        \centering
        \includegraphics[scale=0.7]{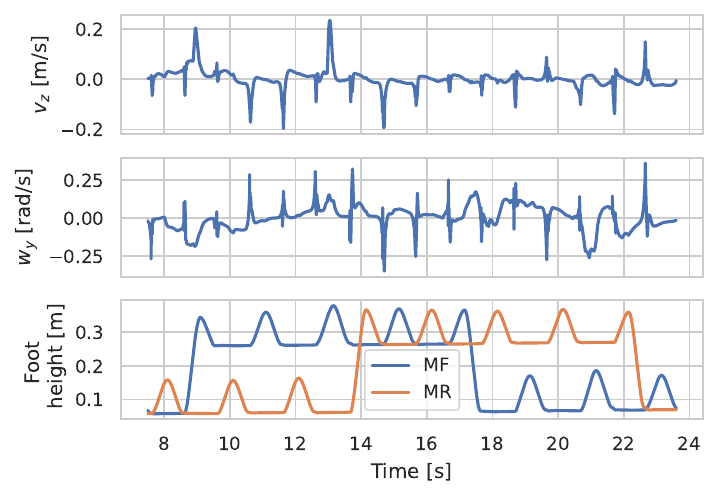}
        \subcaption{NMPC + CBF-QP}
        \vspace{2mm}
    \end{minipage}
    \begin{minipage}{1.0 \linewidth}
        \centering
        \includegraphics[scale=0.7]{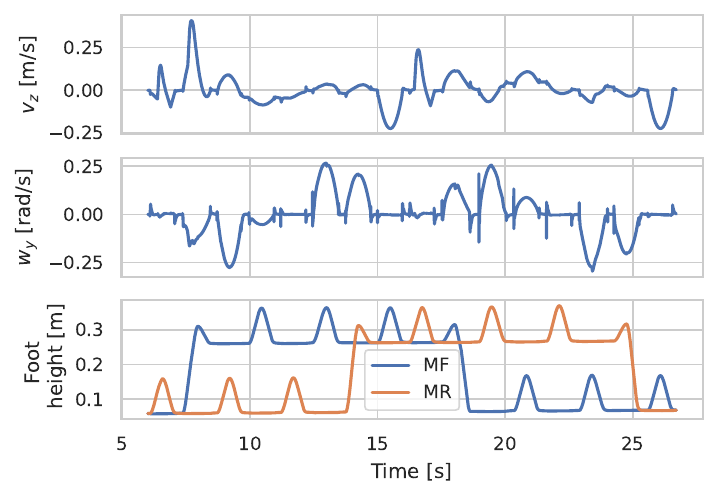}
        \subcaption{CBF-QP}
    \end{minipage}
  \caption{Time histories of the step traversing simulations by the proposed NMPC and previous CBF-QP controller. $v_z$ is the base local $z$ linear velocity, $w_y$ is the base local angular velocity around $y$ axis.
  The MF foot clibmed up the step first and MR foot clibmed down from it last, which indicates the start and end of the step traverse.
  }
  \label{fig:simComparison}
\end{figure}

\begin{table}[tb]
  \captionsetup{width=1.0\linewidth}
  \caption{Comparison between the proposed NMPC and previous controller in the step traversing simulations.}
  \label{table:simComparison}
  \centering
  \begin{tabular}{|c|c|c|}
    \hline
    & NMPC + CBF-QP & CBF-QP \\
    \hline
    Traverse time [s] & 14 & 18 \\
    Average $|v_z|$ & 0.021 & 0.055 \\
    Average $|w_y|$ & 0.056 & 0.065 \\
    \hline
  \end{tabular}
\end{table}

\subsection{Simulation Comparison}\label{Sec:experimentsResultsSim}
We compared the proposed framework combining the NMPC and CBF-QP and the previous controller using only CBF-QP (\cite{takasugi2023realtime}) through simulations in which the robot traverses a 20 cm step by trotting.
Both methods could successfully traverse the step with a trotting gait while producing in different performances.
Fig. \ref{fig:simComparison} shows the time histories of base local $v_z$ (base local $z$ linear velocity), base local $w_y$ (base local angular velocity around $y$ axis), and MF and MR feet heights.
Velocities $v_z$ and $w_y$ are preferably small, but inevitably change along with the terrain.
Table \ref{table:simComparison} summarizes the averages of $|v_z|$ and average $|w_y|$.
We can see that NMPC reduced the unnecessary body velocities $v_z$ and $w_y$ compared with the previous method.
The table also shows that the proposed method achieved a faster traverse time than the previous controller.

\begin{figure}[tb]
    \centering
    \centering
    \includegraphics[scale=0.72]{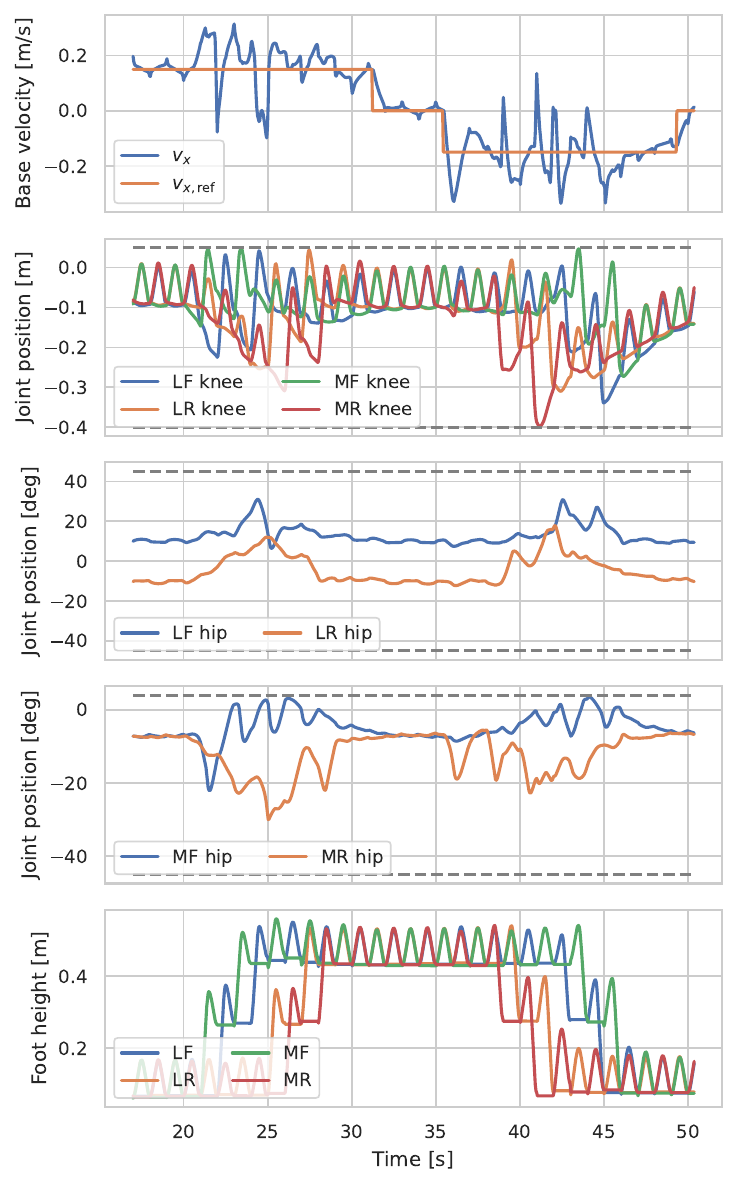}
    \caption{Time histories of the base linear velocity, joint positions, and feet heights of the hardware experiment. RF and RR legs behaved almost the same as LF and LR legs. The dotted gray lines show the joint position limits.}
  \label{fig:ex3MpcJointPosition}
\end{figure}

\begin{figure*}
  \centering
  \includegraphics[scale=0.55]{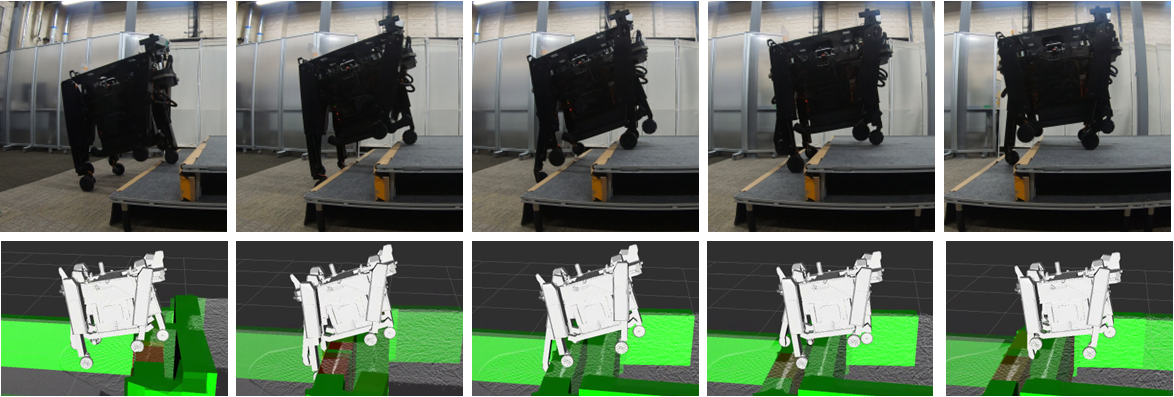}
  \caption{Top row: Snapshots of Tachyon 3 ascending two steps in the hardware experiment. Bottom row: Perception results visualized on rviz corresponding to the upper snapshots. In the lower figures, the white dots show the accumulation point cloud and green and red boxes show the obstacles due to steps.}
  \label{fig:snapshot2Steps}
\end{figure*}

\begin{figure}[tb]
    \centering
    \centering
    \includegraphics[scale=0.72]{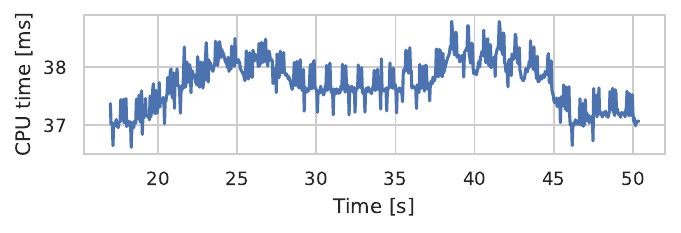}
    \caption{CPU time of each NMPC update of the hardware experiment.}
  \label{fig:ex3MpcCpuTime}
\end{figure}

\subsection{Hardware Experiments}\label{Sec:experimentsResults}

In hardware experiment, Tachyon 3 went back and forth over two steps (ascending and descending 20 cm and 16.5 cm steps) via the proposed NMPC.
Fig. \ref{fig:ex3MpcJointPosition} shows the time histories of base linear velocity, joint positions, and feet heights.
From Fig. \ref{fig:ex3MpcJointPosition}, we observe that the joint positions were close to the joint limit boundaries that are showed as dotted gray lines, which shows that Tachyon 3 required the severe motion generation close to constraint boundaries.
Fig. \ref{fig:snapshot2Steps} shows snapshots of the experiment.
We can see that the feet have been moving close to collisions with the environment, which also shows that the motion is close to constraint boundaries.
We also observed oscillation of the steps themselves which worked as a disturbance.
Nevertheless, the proposed NMPC could successfully traverse these steps with the online optimization subject to the physical and environmental constraints retrieved from on-the-fly perception.
Fig. \ref{fig:ex3MpcCpuTime} shows the CPU time of each NMPC update and we can see that the proposed NMPC run up to 25 Hz on the on-board PC.

\section{Conclusion}\label{Sec:Conclusion}
In this paper, we have presented an NMPC framework toward versatile motion generation for telescopic-wheeled-legged robot Tachyon 3, the unique hardware structure of which poses challenges in control and motion planning.
We have applied the full-centroidal NMPC formulation that can capture accurate kinematics constraints and dynamics of Tachyon 3 with a moderate computational time.
We have developed control pipelines using the internal state integrator to apply the NMPC to Tachyon 3, the actuators of which employ high-gain position-controllers.
We conducted simulation and hardware experiments on the perceptive locomotion of Tachyon 3 over structured terrains.
The simulation study showed that the proposed NMPC performs better than our previous controller.
The hardware experiment demonstrated that the proposed method can generate smooth and dynamic motion under harsh physical and environmental constraints in the real world.

Our future work is to explore the capabilities of the proposed NMPC in a wider variety of settings, such as narrow spaces, cluttered environments, and curved surfaces.

\bibliography{ifacconf}

\begin{thebibliography}{24}
\providecommand{\natexlab}[1]{#1}
\providecommand{\url}[1]{\texttt{#1}}
\providecommand{\urlprefix}{URL }
\expandafter\ifx\csname urlstyle\endcsname\relax
  \providecommand{\doi}[1]{doi:\discretionary{}{}{}#1}\else
  \providecommand{\doi}{doi:\discretionary{}{}{}\begingroup
  \urlstyle{rm}\Url}\fi

\bibitem[{Bjelonic et~al.(2022)Bjelonic, Grandia, Geilinger, Harley, Medeiros,
  Pajovic, Jelavic, Coros, and Hutter}]{bjelonic2022offline}
Bjelonic, M., Grandia, R., Geilinger, M., Harley, O., Medeiros, V.S., Pajovic,
  V., Jelavic, E., Coros, S., and Hutter, M. (2022).
\newblock Offline motion libraries and online mpc for advanced mobility skills.
\newblock \emph{The International Journal of Robotics Research}, 41(9-10),
  903--924.

\bibitem[{Bjelonic et~al.(2021)Bjelonic, Grandia, Harley, Galliard, Zimmermann,
  and Hutter}]{bjelonic2021whole}
Bjelonic, M., Grandia, R., Harley, O., Galliard, C., Zimmermann, S., and
  Hutter, M. (2021).
\newblock Whole-body {MPC} and online gait sequence generation for
  wheeled-legged robots.
\newblock In \emph{2021 IEEE/RSJ International Conference on Intelligent Robots
  and Systems (IROS)}, 8388--8395. IEEE.

\bibitem[{Bock and Plitt(1984)}]{bock1984multiple}
Bock, H.G. and Plitt, K.J. (1984).
\newblock A multiple shooting algorithm for direct solution of optimal control
  problems.
\newblock \emph{IFAC Proceedings Volumes}, 17(2), 1603--1608.

\bibitem[{Carpentier and Mansard(2018)}]{carpentier2018analytical}
Carpentier, J. and Mansard, N. (2018).
\newblock Analytical derivatives of rigid body dynamics algorithms.
\newblock In \emph{Robotics: Science and systems (RSS 2018)}.

\bibitem[{Carpentier et~al.(2019)Carpentier, Saurel, Buondonno, Mirabel,
  Lamiraux, Stasse, and Mansard}]{carpentier2019pinocchio}
Carpentier, J., Saurel, G., Buondonno, G., Mirabel, J., Lamiraux, F., Stasse,
  O., and Mansard, N. (2019).
\newblock The {P}inocchio {C}++ library: A fast and flexible implementation of
  rigid body dynamics algorithms and their analytical derivatives.
\newblock In \emph{2019 IEEE/SICE International Symposium on System Integration
  (SII)}, 614--619. IEEE.

\bibitem[{Di~Carlo et~al.(2018)Di~Carlo, Wensing, Katz, Bledt, and
  Kim}]{di2018dynamic}
Di~Carlo, J., Wensing, P.M., Katz, B., Bledt, G., and Kim, S. (2018).
\newblock Dynamic locomotion in the {MIT} {C}heetah 3 through convex
  model-predictive control.
\newblock In \emph{2018 IEEE/RSJ International Conference on Intelligent Robots
  and Systems (IROS)}, 1--9. IEEE.

\bibitem[{Diehl et~al.(2005)Diehl, Bock, and Schl{\"o}der}]{diehl2005real}
Diehl, M., Bock, H.G., and Schl{\"o}der, J.P. (2005).
\newblock A real-time iteration scheme for nonlinear optimization in optimal
  feedback control.
\newblock \emph{SIAM Journal on control and optimization}, 43(5), 1714--1736.

\bibitem[{Farshidian et~al.(2017)Farshidian, Neunert, Winkler, Rey, and
  Buchli}]{farshidian2017efficient}
Farshidian, F., Neunert, M., Winkler, A.W., Rey, G., and Buchli, J. (2017).
\newblock An efficient optimal planning and control framework for quadrupedal
  locomotion.
\newblock In \emph{2017 IEEE International Conference on Robotics and
  Automation (ICRA)}, 93--100. IEEE.

\bibitem[{Farshidian et~al.(2017--2023)}]{OCS2Farbod}
Farshidian, F. et~al. (2017--2023).
\newblock {OCS2}: An open source library for optimal control of switched
  systems.
\newblock [Online]. Available: \url{https://github.com/leggedrobotics/ocs2}.

\bibitem[{Feller and Ebenbauer(2016)}]{feller2016relaxed}
Feller, C. and Ebenbauer, C. (2016).
\newblock Relaxed logarithmic barrier function based model predictive control
  of linear systems.
\newblock \emph{IEEE Transactions on Automatic Control}, 62(3), 1223--1238.

\bibitem[{Frison and Diehl(2020)}]{frison2020hpipm}
Frison, G. and Diehl, M. (2020).
\newblock {HPIPM}: a high-performance quadratic programming framework for model
  predictive control.
\newblock \emph{IFAC-PapersOnLine}, 53(2), 6563--6569.

\bibitem[{Grandia et~al.(2023)Grandia, Jenelten, Yang, Farshidian, and
  Hutter}]{grandia2023perceptive}
Grandia, R., Jenelten, F., Yang, S., Farshidian, F., and Hutter, M. (2023).
\newblock Perceptive locomotion through nonlinear model-predictive control.
\newblock \emph{IEEE Transactions on Robotics}.

\bibitem[{Hartley et~al.(2020)Hartley, Ghaffari, Eustice, and
  Grizzle}]{hartley2020contact}
Hartley, R., Ghaffari, M., Eustice, R.M., and Grizzle, J.W. (2020).
\newblock Contact-aided invariant extended kalman filtering for robot state
  estimation.
\newblock \emph{The International Journal of Robotics Research}, 39(4),
  402--430.

\bibitem[{Kajita et~al.(2003)Kajita, Kanehiro, Kaneko, Fujiwara, Harada, Yokoi,
  and Hirukawa}]{kajita2003biped}
Kajita, S., Kanehiro, F., Kaneko, K., Fujiwara, K., Harada, K., Yokoi, K., and
  Hirukawa, H. (2003).
\newblock Biped walking pattern generation by using preview control of
  zero-moment point.
\newblock In \emph{2003 IEEE International Conference on Robotics and
  Automation}, 1620--1626. IEEE.

\bibitem[{Katayama and Ohtsuka(2022)}]{katayama2022whole}
Katayama, S. and Ohtsuka, T. (2022).
\newblock Whole-body model predictive control with rigid contacts via online
  switching time optimization.
\newblock In \emph{2022 IEEE/RSJ International Conference on Intelligent Robots
  and Systems (IROS)}, 8858--8865. IEEE.

\bibitem[{Koenemann et~al.(2015)Koenemann, Del~Prete, Tassa, Todorov, Stasse,
  Bennewitz, and Mansard}]{koenemann2015whole}
Koenemann, J., Del~Prete, A., Tassa, Y., Todorov, E., Stasse, O., Bennewitz,
  M., and Mansard, N. (2015).
\newblock Whole-body model-predictive control applied to the hrp-2 humanoid.
\newblock In \emph{2015 IEEE/RSJ International Conference on Intelligent Robots
  and Systems (IROS)}, 3346--3351. IEEE.

\bibitem[{Kuindersma et~al.(2016)Kuindersma, Deits, Fallon, Valenzuela, Dai,
  Permenter, Koolen, Marion, and Tedrake}]{kuindersma2016optimization}
Kuindersma, S., Deits, R., Fallon, M., Valenzuela, A., Dai, H., Permenter, F.,
  Koolen, T., Marion, P., and Tedrake, R. (2016).
\newblock Optimization-based locomotion planning, estimation, and control
  design for the atlas humanoid robot.
\newblock \emph{Autonomous robots}, 40, 429--455.

\bibitem[{Mastalli et~al.(2022)Mastalli, Merkt, Xin, Shim, Mistry, Havoutis,
  and Vijayakumar}]{mastalli2022agile}
Mastalli, C., Merkt, W., Xin, G., Shim, J., Mistry, M., Havoutis, I., and
  Vijayakumar, S. (2022).
\newblock Agile maneuvers in legged robots: A predictive control approach.
\newblock \emph{arXiv preprint arXiv:2203.07554}.

\bibitem[{Nocedal and Wright(1999)}]{nocedal1999numerical}
Nocedal, J. and Wright, S.J. (1999).
\newblock \emph{Numerical optimization}.
\newblock Springer.

\bibitem[{Orin et~al.(2013)Orin, Goswami, and Lee}]{orin2013centroidal}
Orin, D.E., Goswami, A., and Lee, S.H. (2013).
\newblock Centroidal dynamics of a humanoid robot.
\newblock \emph{Autonomous Robots}, 35, 161--176.

\bibitem[{Rawlings et~al.(2017)Rawlings, Mayne, and Diehl}]{rawlingsmodel}
Rawlings, J.B., Mayne, D.Q., and Diehl, M.M. (2017).
\newblock Model predictive control: Theory, computation, and design.

\bibitem[{Sleiman et~al.(2021)Sleiman, Farshidian, Minniti, and
  Hutter}]{sleiman2021unified}
Sleiman, J.P., Farshidian, F., Minniti, M.V., and Hutter, M. (2021).
\newblock A unified {MPC} framework for whole-body dynamic locomotion and
  manipulation.
\newblock \emph{IEEE Robotics and Automation Letters}, 6(3), 4688--4695.

\bibitem[{Takasugi et~al.(2023)Takasugi, Kinoshita, Kamikawa, Tsuzaki,
  Sakamoto, Kai, and Kawanami}]{takasugi2023realtime}
Takasugi, N., Kinoshita, M., Kamikawa, Y., Tsuzaki, R., Sakamoto, A., Kai, T.,
  and Kawanami, Y. (2023).
\newblock Real-time perceptive motion control using control barrier functions
  with analytical smoothing for six-wheeled-telescopic-legged robot {T}achyon
  3.
\newblock \emph{arXiv preprint arXiv:2310.11792}.

\bibitem[{W{\"a}chter and Biegler(2006)}]{wachter2006implementation}
W{\"a}chter, A. and Biegler, L.T. (2006).
\newblock On the implementation of an interior-point filter line-search
  algorithm for large-scale nonlinear programming.
\newblock \emph{Mathematical programming}, 106, 25--57.

\end{thebibliography}

\end{document}